\documentclass{article}

\usepackage[nonatbib,final]{neurips_2019}

\usepackage[utf8]{inputenc} 
\usepackage[T1]{fontenc}    
\usepackage{url}            
\usepackage{amsmath}
\usepackage{booktabs}       
\usepackage{amsfonts}       
\usepackage{nicefrac}       
\usepackage{microtype}      
\usepackage{graphicx}
\usepackage{caption,stackengine}

\title{Formatting Instructions For NeurIPS 2019}

\title{Neural Network Surgery with Sets}
\author{
Jonathan Raiman\thanks{equal contribution} \\
Dali\\
\texttt{jonathan@dali.ml} \\
\And
Susan Zhang\footnotemark[1] \\
OpenAI\\
\texttt{susan@openai.com}
\And
Christy Dennison\thanks{involved in development, debugging, profiling, and testing of the first prototype of this work.} \\
OpenAI\\
\texttt{christy@openai.com}
}

\begin{document}

\maketitle
\begin{abstract}
The cost to train machine learning models has been increasing exponentially \cite{computepost}, making exploration and research into the correct features and architecture a costly or intractable endeavor at scale.
However, using a technique named ``surgery'' OpenAI Five was continuously trained to play the game DotA 2 over the course of 10 months through 20 major changes in features and architecture.
Surgery transfers trained weights from one network to another after a selection process to determine which sections of the model are unchanged and which must be re-initialized.
In the past, the selection process relied on heuristics, manual labor, or pre-existing boundaries in the structure of the model, limiting the ability to salvage experiments after modifications of the feature set or input reorderings.

We propose a solution to automatically determine which components of a neural network model should be salvaged and which require retraining. We achieve this by allowing the model to operate over discrete sets of features and use set-based operations to determine the exact relationship between inputs and outputs, and how they change across tweaks in model architecture. In this paper, we introduce the methodology for enabling neural networks to operate on sets, derive two methods for detecting feature-parameter interaction maps, and show their equivalence. We empirically validate that we can surgery weights across feature and architecture changes to the OpenAI Five model.
\end{abstract}

\section{Introduction}

The computational cost of training neural networks has been shown to be growing at an exponential rate \cite{computepost}. The ability to repurpose or fine-tune machine learning models for different tasks has in parallel been touted as an approach to reap the benefits of a costly investment in a high capacity model. Starting with word vectors \cite{mikolov2013distributed}, base layers of computer vision models, and lately with generative models, Transformers, and reinforcement learning agents, larger and larger components from a trained model can be used on a new task or modified to satisfy new requirements, and attempts have been made to understand how best to achieve this.  For instance, \cite{alain2016understanding} used linear classifier probes to try and determine the utility of layers within a network, while \cite{chen2015net2net} looked at which portions of a model can be transferred.  In simpler cases, it has been found that the output of a network is sufficiently similar to an existing task, and thus the majority of the network can be kept as is \cite{johnson2016densecap,kitaev2018multilingual,raiman2018deeptype,houlsby2019parameter}.

\begin{figure}
\centering
\begin{minipage}[t]{.49\textwidth}
\centering
\includegraphics[width=1\linewidth]{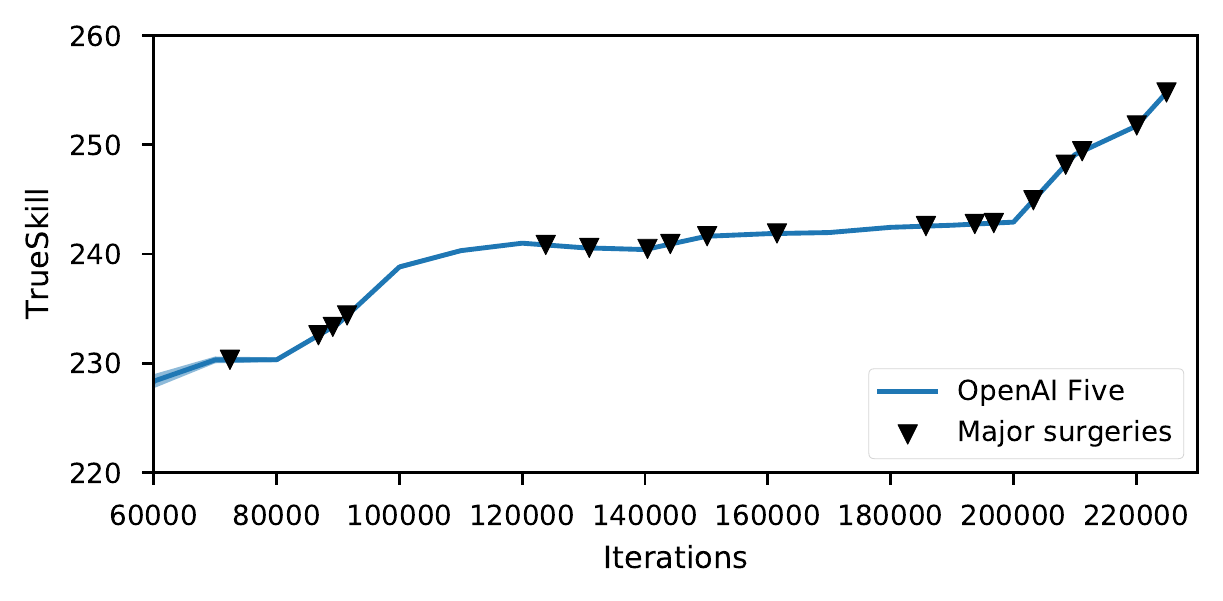}
\caption{Over the course of training, OpenAI Five went through 18 major surgeries. Surgery events are marked on the skill curve.
We measure progress using matches between the training model and reference opponents to compute a TrueSkill \cite{herbrich2007trueskill} rating. \label{figure:trueskill}
}

\end{minipage}
\hfill
\begin{minipage}[t]{.49\textwidth}
    \centering
\includegraphics[width=1\linewidth]{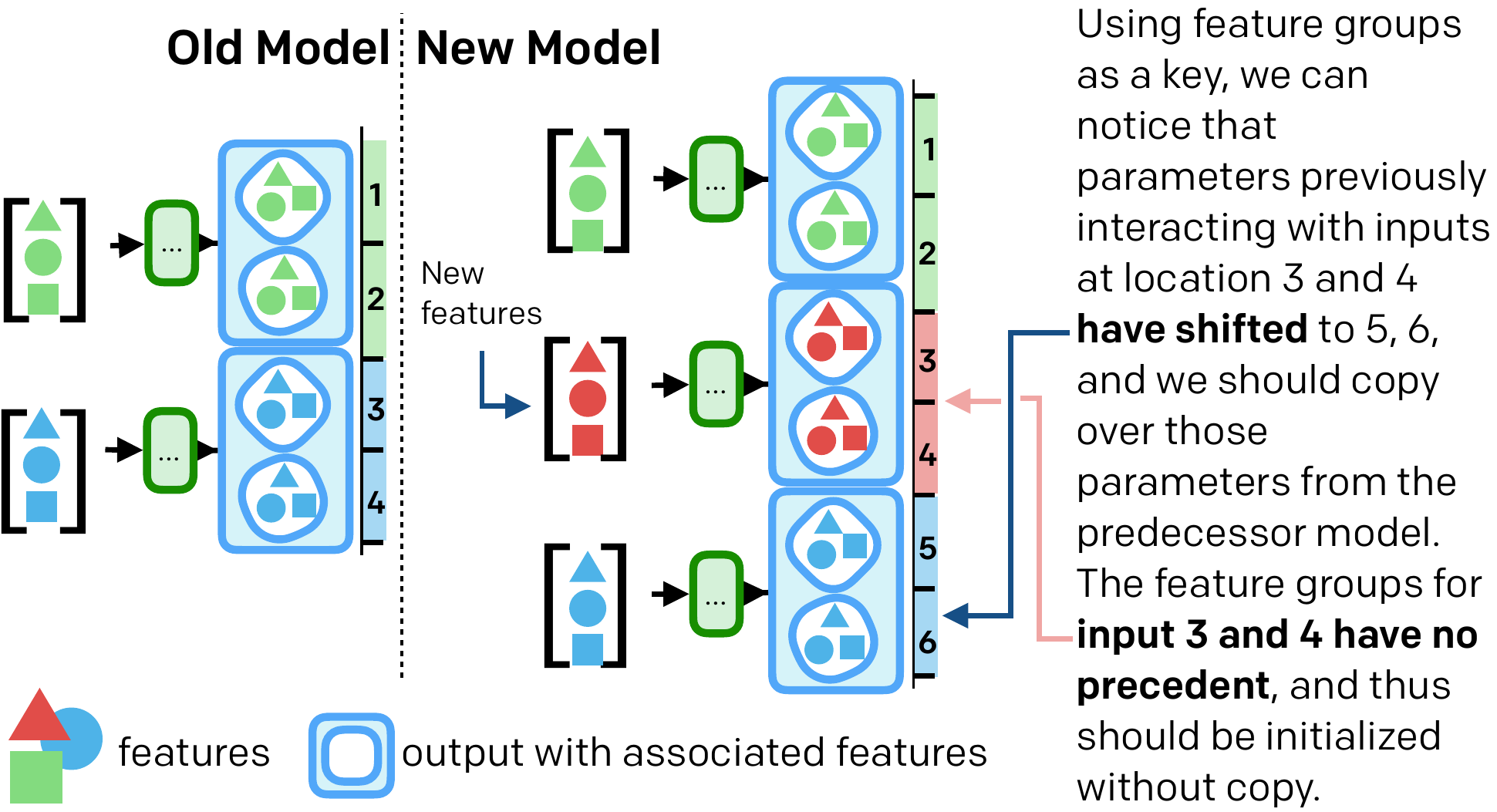}
\caption{When comparing feature-parameter interaction maps across an old and new model, we can notice the introduction of new features and whether features that exist across both generations have shifted places. \label{figure:newfeatures}}
\end{minipage}
\end{figure}

While some transfer cases have natural boundaries along which parameters can be transferred, we find that domains with semantically rich and complex input features pose an important difficulty to overcome when attempting to gradually incorporate them into a growing model.
In the development of OpenAI Five\footnote{OpenAI Five is a deep reinforcement learning agent trained using selfplay that defeated the world champions OG at the game DotA 2.} \cite{OpenAI_dota}, the input domain would periodically change to incorporate new heroes or additional observable features that were deemed necessary to master the game across changes in game versions, as depicted in Figure~\ref{figure:trueskill}.  These changes would pose significant challenges for continued training of the model, motivating the development and formalization of a methodology named ``surgery'' for automatically discovering which parameters to port over.


The main contributions of this work are a formalization of the discovery of surgery steps for porting the parameters from one network to another across feature or architecture changes, two implementations using gradients or boolean logic operations, and empirical validation of surgery applied to feature changes in OpenAI Five.

\section{Approach}

\subsection{Surgery Overview}

For a change to a model $F_{old}$ with parameters $\Theta_{old}$, we would like to define a mapping $\mathcal{M}: \Theta_{old} \to \Theta_{new}$ of model parameters for a new model $F_{new}$ such that
\begin{align}
F_{old}\left(X^{in}_{old}(s);\Theta_{old}\right) = F_{new}\left(X^{in}_{new}(s);\Theta_{new}\right)
\hspace{0.5cm}
\forall \; s \in \mathcal S
\end{align}
where $s$ is a state of the world $\mathcal S$, $X^{in}_{old}(s)$ and $X^{in}_{new}(s)$ are the corresponding featurized inputs to each model at a state $s$. In order to define this mapping $\mathcal{M}$, we require the following:
\begin{enumerate}
    \item An ordered and named schema for all input features in $X^{in}_{old}(\cdot)$ and $X^{in}_{new}(\cdot)$.
    \item A mapping $\Phi_{old}$ from each input feature $x^{in}_i(\cdot) \in X^{in}_{old}(\cdot)$ to each parameter $\theta_j \in \Theta_{old}$ indicating the interaction between the two within the network.  A separate mapping $\Phi_{new}$ is obtained for features in $X^{in}_{new}$ and parameters in $\Theta_{new}$.
    \item The difference between $\Phi_{old}$ and $\Phi_{new}$ is computed. Changes indicate areas where new parameters need to be inserted to preserve output equivalence between $F_{old}$ and $F_{new}$.
\end{enumerate}

\subsection{Input feature schema}
Let $X^{in}(s)$ denote the input features for a given state $s$ that is fed into some model $F$, which takes the form $X^{in}(s) = [x_{\text{max health}}(s), x_{\text{attack damage}}(s), \dots]$.  We would then record an index mapping of: $
\{\text{\small max health} \to 0,\,\, \text{\small attack damage} \to 1, \dots \}.$
We call this snapshot of feature identifiers to their corresponding indices within the model input $X^{in}$ an \textbf{input feature schema}. For notational simplicity we will omit $s$ from $X^{in}$ past this point.

\subsection{Input feature to parameter mapping}

Let $\Theta=\{\theta_i\}_{i\in\mathcal{L}}$ be the set of learnable parameters within a model $F$, where $\mathcal{L}$ is the set of layers within a given model.  We need to define a mapping $\Phi: x^{in}_i \to \theta_j \hspace{0.25cm} \forall \; x^{in}_i \in X^{in}, \theta_j \in \Theta$ which connects input features to the parameters in the model for which these features interact with.  Binary operations between inputs\footnote{Note that we distinguish between \textbf{inputs} and \textbf{input features}. \textbf{Inputs} are used to reference inputs to any layer in the network, including hidden states, whereas \textbf{input features} refers solely to observations directly taken from the world $\mathcal{S}$ and fed into the network initially.} and parameters are defined as ``interactions'', and we can derive this interaction mapping using either gradients or boolean logic.

\subsubsection{Gradient mapping}

One method of obtaining $\Phi$ is to look for cases where parameters receive a nonzero gradient. 
For example, in the case of a single fully-connected (FC) layer with an element-wise differentiable activation function $f$, let us consider the $N\times M$ matrix of weights $W$, a $1\times M$ vector of biases $b$, and a $1\times N$ input vector $X$. We now have $Y=f(X \cdot W + b)$, where $Y$ is the output of this layer. If we have a cost function $C$ that takes $Y$ as its input: $C(Y)=C\left(y_1, y_2, \dots, y_M\right)$, then the gradient of the cost function with respect to the weights of this layer is:
\begin{align} \label{eq:costpartial}
    \frac{\partial\,C(Y)}{\partial W} &= 
    \sum_{k=1}^M\left(\frac{\partial C}{\partial y_k}\frac{\partial y_k}{\partial W}\right)
\end{align}
and since we have
\begin{align}
    \frac{\partial y_k}{\partial W} &= 
    \begin{bmatrix} 
    \frac{\partial y_k}{\partial w_{1,1}} &
    \cdots &
    \frac{\partial y_k}{\partial w_{1,M}}\\
    \vdots & \ddots & \vdots \\
    \frac{\partial y_k}{\partial w_{N,1}} &
    \cdots & 
    \frac{\partial y_k}{\partial w_{N,M}}
    \end{bmatrix}
\end{align}
where $y_k=f\left(\sum_{i=1}^N x_i \cdot w_{i,k}\right)$, the matrix of partials $\frac{\partial y_k}{\partial W}$ reduces down to all zeros except for the $k$-th column. Furthermore, if we were to set the input vector $X$ to be all zeroes except for a 1 at index $t$, the matrix of partials $\frac{\partial y_k}{\partial W}$ would consist of simply a single nonzero value at index $(t,k)$.  Thus from feeding in inputs $X_t$ that is nonzero only at index $t$, and computing the gradient, we can back out all the elements in $W$ that need to be modified if an element of $X$ was changed, since the nonzero values would all reside on row $t$ of $\frac{\partial C(Y)}{\partial W}$.

Using gradients to define $\Phi$ is practical as it relies on existing auto-differentiation tools \cite{abadi2016tensorflow} to collect information. However, poor initialization and saturating functions can mask interactions. We resolve these issues in practice by initializing (non-bias) weights to be strictly positive, normalizing the inputs into sigmoid/tanh to avoid saturation, and replacing $max$-$pool$ by $avg$-$pool$.

If we were to have more than two layers, we would not be able to distinguish changes to the weight matrix in the second layer under the assumptions given in our surgery setup.  That is, given positive weight matrices in our network and the assumption that we do not saturate any of our activation functions, we cannot construct inputs to the second layer that are nonzero at a single index without solving for complex systems of equations.  As a result, the gradient of the cost function at the output of the second layer with respect to the weight matrix in the second layer would be nonzero throughout, and we would not be able isolate changes to the second layer parameters as a function of input changes.

While the limitation to a single FC layer may appear to be somewhat restrictive by design, we can still sufficiently cover all cases where input features to the model are added or removed.
All input features into the model will be fed through some layer after which the output dimension is changed either through a $\mathrm{matmul}$ or a $\mathrm{concat}$ with another feature vector.  In the case of $\mathrm{matmul}$s, we have illustrated above that we can isolate the impact of individual inputs through the index of nonzero gradients. In the $\mathrm{concat}$ case, say with the output of two FC layers processed over two sets of inputs, we can similarly isolate changes in the ordering of the $\mathrm{concat}$ arguments by looking at the position of feature groups as shown in Figure~\ref{figure:latentnestedfc}.

\subsubsection{Boolean logic mapping}
\begin{figure}
\centering
\includegraphics[width=0.75\textwidth]{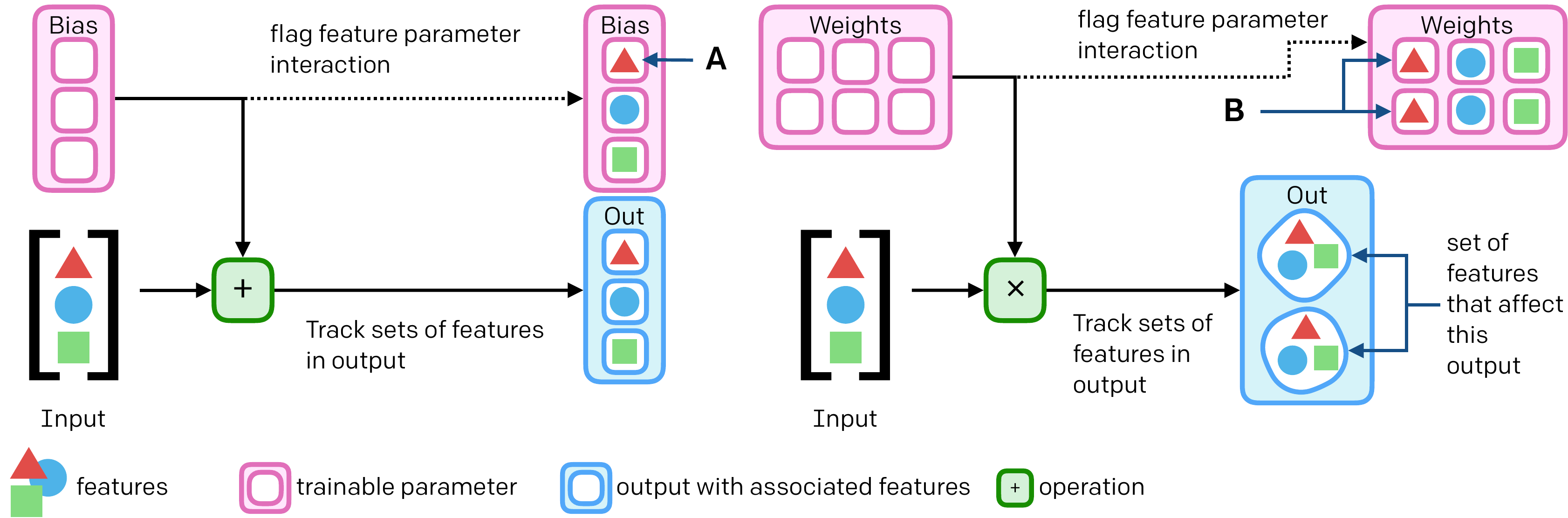}
\captionof{figure}{Neural network is evaluated using boolean operations that track the propagation of features (shown as triangles, circles, and squares) using sets.
}
\label{figure:binary}
\end{figure}
Another method for defining $\Phi$ relies on tracking explicitly when and where a particular input feature is involved in a binary operation with a model parameter. This interaction is stored through a list of boolean flags \footnote{We use bitfield \cite{bitfield} to efficiently store feature sets.} corresponding to each input feature. Every model parameter will maintain its own list of boolean flags, where a true value indicates the parameter's involvement in a binary operation with a given input feature.  An illustration of the propagation and update of these feature flags is shown in Figure~\ref{figure:binary}.  

To create these per-parameter feature flags, we create a copy of the computational graph where the original network operations (matmul, addition, sigmoid, etc...) are implemented using boolean logic rules to track feature interactions between inputs and parameters.  The rules are defined as follows. Let $\mathcal{F}_w$ denote the set of input features $x^{in}_i \in X^{in}$ that interact with a scalar parameter $w$, and let $\mathcal{F}_{x}$ denote the set of input features that feed directly into an input $x$. Using the rules shown in Table~\ref{table:rules}, we can combine feature sets whenever parameters and inputs interact through addition or multiplication. At the end of this process, we obtain a parameter to feature mapping, which we can then do a reverse lookup to create $\Phi$.  This process is also illustrated in Figure~\ref{figure:binary}.

\begin{table}
    \centering
    \begin{tabular}{l|c|c}
    Interaction & Side effects & Output features\\
    \hline
    $w_1 \odot w_2$ & - & $\emptyset$\\
    $w  \odot x \;\text{ or }\; x \odot w$ & $\mathcal{F}_w \to \mathcal{F}_w \cup \mathcal{F}_x$ & $\mathcal{F}_x$\\
    $x_i \odot x_j$ & - & $ \mathcal{F}_{x_i} \cup \mathcal{F}_{x_j}$\\
    \end{tabular}
    \caption{Feature propagation rules in a neural network made of parameters $w$ and inputs $x$. $\odot$ represents an additive or multiplicative interaction between two items.}
    \label{table:rules}
\end{table}

\subsubsection{Gradient and boolean logic mapping equivalence}
\label{section:proof}

We can show that both gradient and boolean logic mappings generate the same $\Phi$ by noticing that the existence of a nonzero gradient for a parameter (when a single input is nonzero) relies solely on the outputs of the forward pass up to this point being positive.  From (\ref{eq:costpartial}), we can see that:
\begin{align}
    \left[\frac{\partial\,C(Y)}{\partial W}\right]_{a,b} = 
    \frac{\partial C}{\partial y_b}\left[\frac{\partial y_b}{\partial W}\right]_{a,b} = 
    \frac{\partial C}{\partial y_b}\cdot \frac{\partial}{\partial w_{a,b}}  f\left(\sum_{i=1}^N x_i \cdot w_{i,b} \right) =  \frac{\partial C}{\partial y_b}\cdot f'\cdot x_a
\end{align}
Since $\frac{\partial C}{\partial y_b}$ is assumed to be nonzero for all $y_b$, and $f'$ is similarly designed to be nonzero, we get:
\begin{align}
    \left[\frac{\partial\,C(Y)}{\partial W}\right]_{a,b} \neq 0 \implies  x_a \neq 0
\end{align}
if there exists $w_{a,b} \in W$ that is multiplied with $x_a$.  It follows that if we were to do a forward pass with an input vector $X$ that is zero except for a 1 at index $t$, any nonzero outputs of $y_k \in Y$ would indicate an interaction between $x_t$ and $w_{t,k}$, which is equivalent to the location of nonzero gradients within $\frac{\partial\,C(Y)}{\partial W}$.

\begin{table}[ht]
    \centering
    \begin{tabular}{l|r|r}
    Interaction & Positive output & Input feature $x^{in}_i$ present in output\\
    \hline
    $x_1 \cdot x_2$ & $x_1 > 0 \land x_2 > 0$ & $(x^{in}_i \in \mathcal{F}_{x_1}) \lor (x^{in}_i \in \mathcal{F}_{x_2})$\\
    $x \cdot w$ & $x > 0$ & $x^{in}_i \in \mathcal{F}_x$\\
    $x_1 + x_2$ & $x_1 > 0 \lor x_2 > 0$ & $(x^{in}_i \in \mathcal{F}_{x_1}) \lor (x^{in}_i \in \mathcal{F}_{x_2})$\\
    $x + w$ & $x > 0$ & $x^{in}_i \in \mathcal{F}_x$\\
    \end{tabular}
    \caption{Boolean logic condition rules for propagating features when evaluating a neural network made of parameters $w$ and inputs $x$.}
    \label{table:positivity}
\end{table}

In order for $x_a$ to be positive it must be either an input feature that was set to 1, or the output of some other layer in the network. In Table~\ref{table:positivity} we describe how a positive output relies either on any input being positive (in the case of matmuls), or a positive input element at the same index for element-wise operations. We can also observe that the truth table under the positive output conditional is identical to the truth table from conditional yielding the presence of a feature $x^{in}_i$ in the output\footnote{
Except in the case of the product of two inputs $x_1$ and $x_2$ which gradient mapping cannot trace if $x_1$ and $x_2$ are inputs with mutually exclusive features. However, we did not encounter this case in practice.}, and thus propagation of positivity and the propagation of features operate identically. Gradient and boolean logic mappings are therefore equivalent, since they both only rely on detecting a positive value at $x_a$, and the operations that determine whether $x_a$ is positive are identical.

\subsection{Mapping Differences}

The final stage for obtaining surgery steps is to compare $\Phi_{old}$ with $\Phi_{new}$. We construct a lookup table using $\Phi$ that is keyed by feature groups and maps to parameters that interact with them\footnote{See boxes A and B in Figure~\ref{figure:binary}, where the triangle feature refers to specific parameters in the bias or weights matrix}. Whenever new features are introduced, they will alter the feature groups so that $\Phi_{old}$ and $\Phi_{new}$ will not share this key in their lookup tables. We can use this mismatch to identify parameters that need to be reinitialized.

When new sets of features are introduced and concatenated with an existing input, the existing feature group keys will make it obvious that features have shifted location; this location shift information is useful for finding the parameters that should be copied from the old model. Figure~\ref{figure:newfeatures} is an illustration new feature introduction and shift detection.

\begin{figure}
\centering
\includegraphics[width=0.84\textwidth]{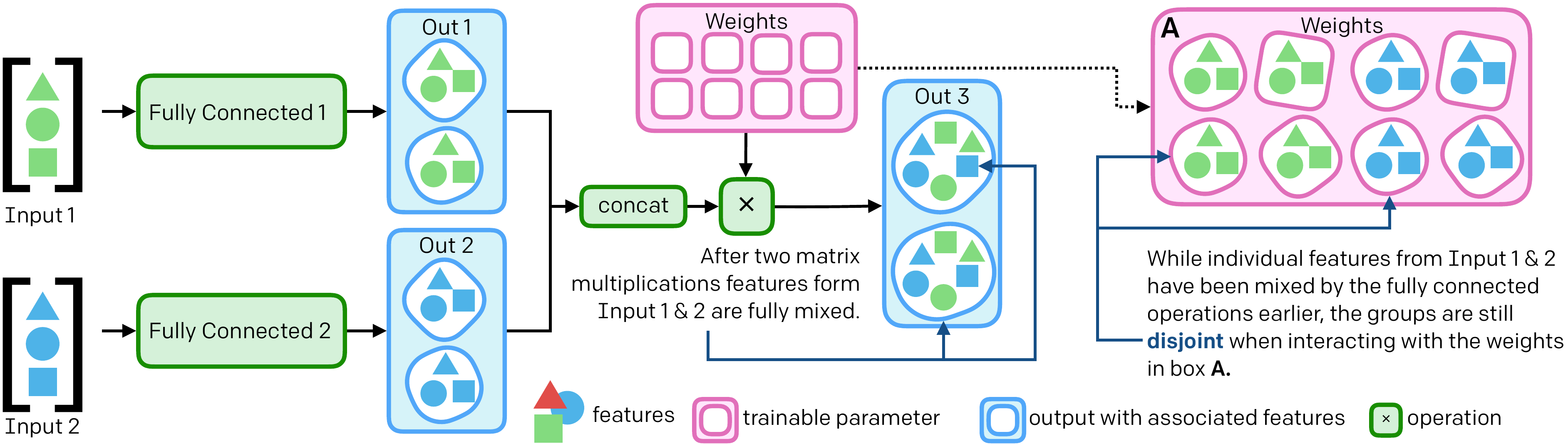}
\caption{The interaction between features and parameters can be tracked across one matmul or FC layer, but individual features will then be fully mixed. However, when concatenating groups of non-overlapping features as input to a deeper FC layer, the interaction between feature groups can still be distinguished (see box A).}
\label{figure:latentnestedfc}
\end{figure}

However, after one FC layer or $\mathrm{matmul}$, features can become mixed (see Out 1 and 2 in Figure~\ref{figure:latentnestedfc}). making it impossible to track propagation past this point using purely input features. Nonetheless, it is possible to track the movement of feature groups that have been processed in isolation. For instance, in the case of OpenAI Five, information about allied and enemy units was processed by separate network components, and only later concatenated and fed to the main LSTM \cite{OpenAI_dota}. This architectural pattern is described pictorially in Figure~\ref{figure:latentnestedfc}, where Input 1 and 2 go through separate FC layers. Their outputs are later concatenated and multiplied by the same matrix (box A), and we can see how feature group movement, insertion, or deletion is visible in the interaction with this deeper weight matrix.

\section{Results}
To test whether we can use gradient or boolean logic mapping to automatically obtain surgery steps for transferrings parameters from one model version to the next, we add a group of 12 per-hero features 
and verify that we are able to preserve most of the trained weights when we use proper initialization and avoidance of zero-gradient functions.

We conduct our experiments on the model that was used during the OG-OpenAI Five match. In Table~\ref{table:correctness} we report the percentage of total feature-parameter interactions detected, as well as the percentage of the original model's parameters that can be transferred to the new model under different initializations and zero-gradient function replacements. We find that random positive initialization and zero-gradient function replacement achieves the highest number of transferred parameters, and detects all feature-parameter interactions. Random initialization on its own masks interactions (49.42\%) and prevents the same degree of transfer.

Unlike gradient mapping, boolean logic mapping has the desirable trait that it is agnostic to initialization or the presence of zero-gradient functions.
In Section~\ref{section:proof} we established equivalence between gradient and boolean logic mapping, and in Table~\ref{table:correctness} we empirically verify this holds and observe the the same number of parameters are transferred by both techniques.

A further advantage of boolean logic mapping is the ability to trace all feature-parameter interactions simultaneously, which we hypothesize should accelerate the map generation process; we measure time taken by both techniques and notice boolean logic mapping is $\approx 24$ times faster (Table~\ref{table:correctness}).

The underlying motivation for surgery is the ability to amortize training time while still being able to introduce new features. We find that surgery does indeed reduce the expected overall time to attain professional level play at DotA 2. 
Specifically, training from scratch (Rerun in \cite{OpenAI_dota2019}) took approximately 2 months, therefore retraining for each of the 20 major surgeries shown in Figure~\ref{figure:trueskill} would have taken 40 months, while training with surgery completed in only 10 months.




\begin{table}
    \centering
    \begin{tabular}{lrrrr}
    \small Mapping & \parbox[c]{1.7cm}{\small Interactions\\Found\strut} & \parbox[c]{1.7cm}{\small  Params\\Transferred\strut} & \small Time (secs, $\mu \pm \sigma$)\\
    \hline
    \hline
    {\small Gradient (random $+$ init.)} & 0.52\% & 90.13\% & -\\
    {\small Gradient (random init.)} & 49.42\% & 94.91\% & -\\
    {\small Gradient (random $+$ init., $\max \to \mathrm{avg}$)} & {\bf 100\%} & {\bf 98.68\%} & $907.92 \pm 195.31$\\
    {\small Boolean logic} & {\bf 100\%} & {\bf 98.68\%} & $37.88 \pm 0.76$\\
    \end{tabular}
    \caption{Surgery correctness is sensitive to initialization and elimination of zero-gradient functions. Boolean logic and gradient mapping find the same interactions and  transfer an equal number of parameters across feature changes. Surgery time comparison on a 2.9 GHz Intel Core i7 computer.}
    \label{table:correctness}
\end{table}

\section{Conclusion}
The exponential rise in computation found in machine learning experiments motivates the creation of tools that amortize the cost to train new models. Surgery has emerged as a powerful method for transferring trained weights across model changes and warm starting experiments without having to pay the price of starting from scratch. Knowing which parts of a model should be kept remains a complicated task that requires human labor and expertise.

In this paper, we introduce a methodology for automatically suggesting surgical steps, and present how it was successfully used to continuously train OpenAI Five across 20 surgeries over the course of 10 months. We show how we can evaluate a neural network using sets to trace the interaction between features and parameters deep inside the network. Building upon this technique, we present two different methods to generate these interaction maps and show their equivalence. We empirically validate that we can detect 100\% of the feature-parameter interactions present in the model when we use a positive initialization and remove zero-gradient functions. By observing the difference in the interaction maps across a feature change we can transfer 98.68\% of the parameters automatically.

We believe neural network surgery with sets can become a powerful tool to modify and reuse parameters in long running machine learning experiments. An important area for future work is determining how to robustly continue training surgeried models without scrambling transferred weights.

\section*{Acknowledgement}
We wish to thank the anonymous reviewers for their valuable feedback and remarks, along with Jakub Pachocki and Szymon Sidor for their work on gradient-based surgery.

\bibliographystyle{unsrt}
\bibliography{bibliography}

\begin{thebibliography}{10}

\bibitem{computepost}
Dario Amodei and Danny Hernandez.
\newblock A{I} and compute.
\newblock \url{https://openai.com/blog/ai-and-compute/}, 2018.

\bibitem{mikolov2013distributed}
Tomas Mikolov, Ilya Sutskever, Kai Chen, Greg~S Corrado, and Jeff Dean.
\newblock Distributed representations of words and phrases and their
  compositionality.
\newblock In {\em Advances in neural information processing systems}, pages
  3111--3119, 2013.

\bibitem{alain2016understanding}
Guillaume Alain and Yoshua Bengio.
\newblock Understanding intermediate layers using linear classifier probes.
\newblock {\em arXiv preprint arXiv:1610.01644}, 2016.

\bibitem{chen2015net2net}
Tianqi Chen, Ian Goodfellow, and Jonathon Shlens.
\newblock Net2net: Accelerating learning via knowledge transfer.
\newblock {\em arXiv preprint arXiv:1511.05641}, 2015.

\bibitem{johnson2016densecap}
Justin Johnson, Andrej Karpathy, and Li~Fei-Fei.
\newblock Densecap: Fully convolutional localization networks for dense
  captioning.
\newblock In {\em Proceedings of the IEEE Conference on Computer Vision and
  Pattern Recognition}, pages 4565--4574, 2016.

\bibitem{kitaev2018multilingual}
Nikita Kitaev and Dan Klein.
\newblock Multilingual constituency parsing with self-attention and
  pre-training.
\newblock {\em arXiv preprint arXiv:1812.11760}, 2018.

\bibitem{raiman2018deeptype}
Jonathan~Raphael Raiman and Olivier~Michel Raiman.
\newblock Deeptype: multilingual entity linking by neural type system
  evolution.
\newblock In {\em Thirty-Second AAAI Conference on Artificial Intelligence},
  2018.

\bibitem{houlsby2019parameter}
Neil Houlsby, Andrei Giurgiu, Stanislaw Jastrzebski, Bruna Morrone, Quentin
  De~Laroussilhe, Andrea Gesmundo, Mona Attariyan, and Sylvain Gelly.
\newblock Parameter-efficient transfer learning for nlp.
\newblock {\em arXiv preprint arXiv:1902.00751}, 2019.

\bibitem{OpenAI_dota}
OpenAI.
\newblock Openai {F}ive.
\newblock 2018.

\bibitem{herbrich2007trueskill}
Ralf Herbrich, Tom Minka, and Thore Graepel.
\newblock Trueskill™: a bayesian skill rating system.
\newblock In {\em Advances in neural information processing systems}, pages
  569--576, 2007.

\bibitem{abadi2016tensorflow}
Mart{\'\i}n Abadi, Paul Barham, Jianmin Chen, Zhifeng Chen, Andy Davis, Jeffrey
  Dean, Matthieu Devin, Sanjay Ghemawat, Geoffrey Irving, Michael Isard, et~al.
\newblock Tensorflow: A system for large-scale machine learning.
\newblock In {\em 12th $\{$USENIX$\}$ Symposium on Operating Systems Design and
  Implementation ($\{$OSDI$\}$ 16)}, pages 265--283, 2016.

\bibitem{bitfield}
Steve Stagg.
\newblock bitfield: Python fast integer set implementation.
\newblock \url{https://github.com/stestagg/bitfield}, 2013.

\bibitem{OpenAI_dota2019}
OpenAI.
\newblock Playing {D}ota 2 with {L}arge {S}cale {D}eep {R}einforcement
  {L}earning, 2019.

\end{thebibliography}

\end{document}